\def\year{2021}\relax
\documentclass[letterpaper]{article} 
\usepackage{aaai21}  
\usepackage{times}  
\usepackage{helvet} 
\usepackage{courier}  
\usepackage[hyphens]{url}  
\usepackage{graphicx} 
\usepackage{algorithm}
\usepackage[noend]{algpseudocode}
\usepackage{latexsym}
\usepackage{amsfonts}
\usepackage[hang]{footmisc}
\usepackage{float}
\usepackage{subfigure}
\usepackage{todonotes}
\usepackage[switch]{lineno}

\usepackage{booktabs}
\usepackage{natbib}  
\usepackage{caption} 
\title{A Semi-Supervised Deep Clustering Pipeline for Mining Intentions From Texts}

\author{
    Xinyu Chen and Ian Beaver
    \\
}
\affiliations{

    Verint Systems Inc.\\
    175 Broadhollow Rd, Ste 100\\
    Melville, NY 11747\\
    Xinyu.Chen@verint.com, Ian.Beaver@verint.com

}

\begin{document}
\maketitle

\begin{abstract}
Mining the latent intentions from large volumes of natural language inputs is a key step to help data analysts design and refine Intelligent Virtual Assistants (IVAs) for customer service and sales support.  To aid data analysts in this task we have created Verint Intent Manager (VIM), an analysis platform that combines unsupervised and semi-supervised approaches to help analysts quickly surface and organize relevant user intentions from conversational texts.  For the initial exploration of data we make use of a novel unsupervised and semi-supervised pipeline that integrates the fine-tuning of high performing language models, a distributed k-NN graph building method and community detection techniques for mining the intentions and topics from texts.  The fine-tuning step is not only beneficial but also necessary because pre-trained language models cannot encode texts to efficiently surface particular clustering structures when the target texts are from an unseen domain or the clustering task is not topic detection.  For flexibility we deploy two different clustering approaches: one where the number of clusters must be specified by the analyst and one where the number of clusters is detected automatically with comparable clustering quality but at the expense of additional computation time.  We describe the application and deployment and demonstrate its performance using BERT on three real-world publicly available text mining tasks.  Our experiments show that BERT begins to produce better task-aware representations using a labeled subset as small as $0.5\%$ of the task data.  The clustering quality exceeds the state-of-the-art results when BERT is fine-tuned with labeled subsets of only $2.5\%$ of the task data.  As deployed in the VIM application, this flexible clustering pipeline produces high quality results, improving the performance of data analysts and reducing the time it takes to surface intentions from customer service data, thereby reducing the time it takes to build and deploy IVAs in new domains.
\end{abstract}

\section{Introduction}

Intelligent Virtual Assistants (IVAs), e.g., Amazon's Alexa or Apple's Siri, are becoming more popular in customer service and product support tasks~\cite{ram2018conversational}.  At Verint, the process to design and refine IVAs relies on data analysts who are familiar with specific terminology in a given language domain such as transportation or finance to mine customer service texts for latent user intentions, or \textit{intents}.  An intent is the interpretation of user input that allows the IVA to formulate the 'best' possible response and its detection is typically treated as a supervised classification problem in commercial IVAs~\cite{beaver2020automated}.  It is common for these analysts to receive a large batch of customer service logs from a company who would like to deploy an IVA to help automate some aspect of their customer service.  The analysts then need to mine this text data to surface the most common intents and determine which use cases an IVA would be able to easily automate.  They would then recommend a subset of high value intents for a company-specific IVA implementation to be deployed on a website, mobile application, or phone support line.

\begin{figure*}
\centering
\includegraphics[width=\textwidth]{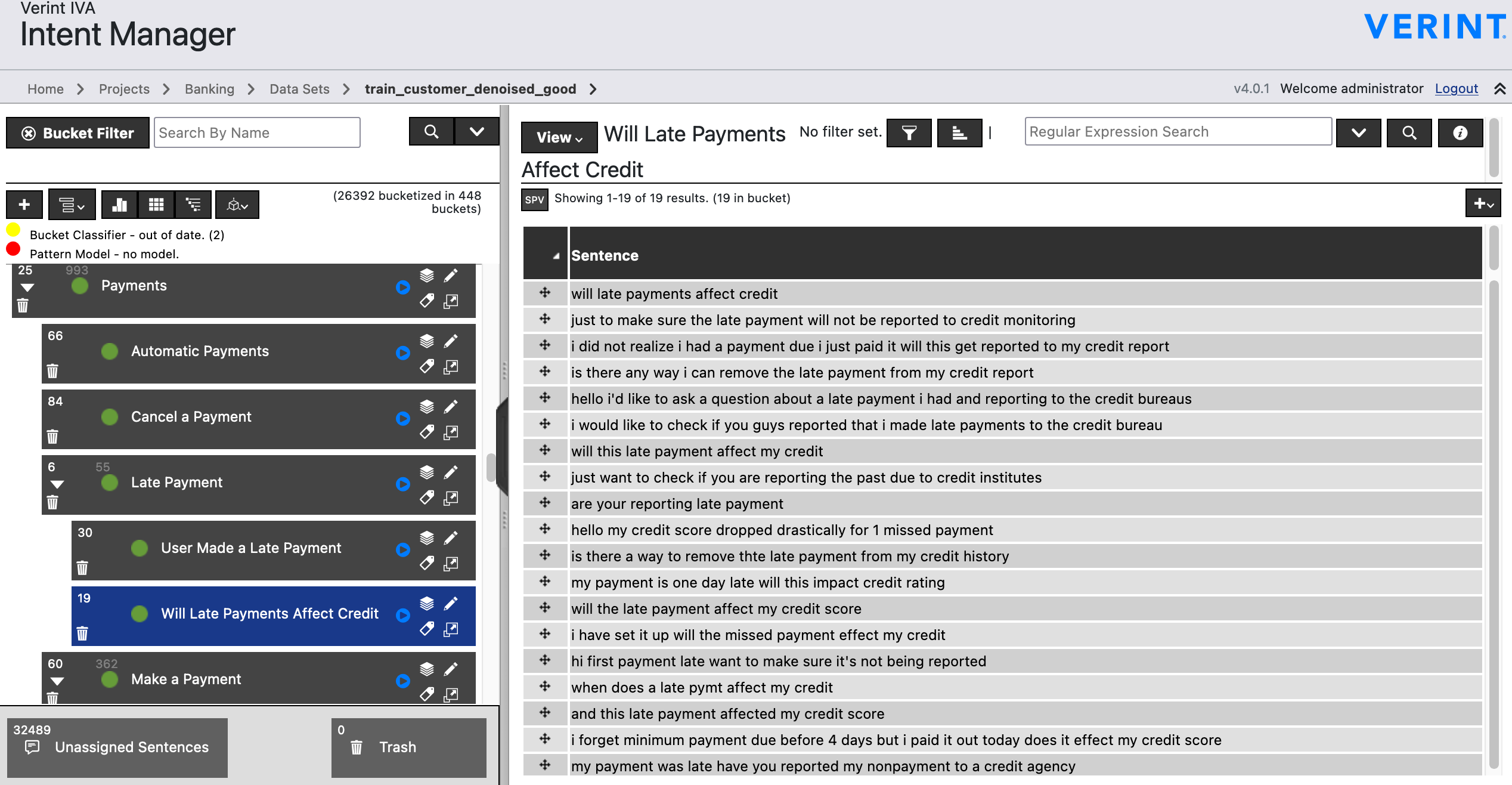}
\caption{ VIM provides an easy drag and drop interface to help analysts review and organize text inputs into hierarchical intents.  The difficulty for analysts is discovering latent intent groupings in large datasets they have little prior experience working with.}
\label{fig:prompt}
\end{figure*}

\textit{Verint Intent Manager (VIM)} is a powerful language classification tool that helps our data analysts to review and organize such large volumes of unlabeled conversational text inputs into various intents. Figure~\ref{fig:prompt} shows the main VIM work space where analysts search, filter, and group unlabeled text inputs (right table) and assign them to labels, then organize the labels into hierarchies (left sidebar)\footnote{A video demonstration of an older (beta) version of this interface is available at https://youtu.be/UjLu1L-ES0w}.  These labeled text inputs are then exported to build intent classifier models for production IVAs. Using text clustering algorithms to provide recommendations for analysts to gain a better understanding of the type of language contained in input texts can partially automate the intent categorization process. However, there are four challenges that affect the performance of text clustering tasks on real-world customer service inputs.

The first challenge is the choice of text data representations. Because the lengths of conversational turns are often short, directly using a binary representation (Bag of Word) or weighted frequencies of words (TF-IDF) for such data leads to the lexical sparsity issue and these representations generally produce poor clustering quality~\cite{aggarwal2012survey, wang2016semi}. We are inspired to encode texts to contextualized word representations with language models~\cite{radford2019language, peters2018deep, devlin2018bert, liu2019roberta} because they greatly improve the performance of Semantic Textual Similarity (STS) tasks~\cite{cer2017semeval}, where the length of STS sentences are close to the length of IVAs' text inputs.

The second challenge is the desired grouping of texts for describing user intentions may be different from those learnt by fully unsupervised approaches~\cite{wang2016semi}. We observed that  language models are pre-trained to maximize the joint probabilities of symbols and words within text sequences. These joint probabilities resemble the definition of \textit{topics} in the topic modeling task~\cite{steyvers2007probabilistic, blei2012probabilistic}, i.e., a topic is a probability distribution over words. Thus we call special attention to the text mining practitioners that unless given further guidance, the contextualized text representations from pre-trained language models are more suited for finding latent topic clusters than surfacing user intentions.

The third challenge is the volume of the input texts. The size of unlabeled conversational logs obtained from contact centers varies from tens of thousands to millions. The performance of the clustering method needs easily scale to such large volumes so that analysts can have the clustering results fast enough to iterate between labeling and evaluation often.

The fourth challenge is the choice of number of clusters by analysts. Often the true number of clusters on a new dataset is unknown and clustering methods that require the true number of clusters may get sub-optimal results when given an inaccurate number of clusters.  We designed VIM to be configurable to use k-means or the Louvain~\cite{blondel2008fast} algorithm depending on whether or not the desired cluster number is known by the analyst beforehand.

To address these challenges for data exploration in the VIM platform,  we proposed and developed an end-to-end text clustering pipeline that integrates four highly successful AI/ML techniques: transfer learning with state-of-the-art language models (e.g., BERT), a high performance library for k-nearest neighbor (k-NN) graphs construction and k-means clustering leveraging multiple GPUs, and a fast community detection algorithm to uncover high quality community structures. The pipeline is flexible and configurable based on the prior knowledge of data. This semi-supervised pipeline makes the following contributions:
\begin{enumerate}
    
    \item We integrated this semi-supervised deep clustering pipeline within the VIM application to facilitate the design and refinement of IVAs. With a small amount of labeled samples, this pipeline adapts the fine-tuned language model to create task-aware text representations and surface intention groups. The deployment of this pipeline greatly improved the usability of VIM for data analysts to review and organize text inputs in production use.
    \item We use real-world text data to showcase the high accuracy of the pipeline in topics or intentions clustering tasks.  Our pipeline greatly exceeds the state-of-art unsupervised and semi-supervised approaches in the Reuters news clustering task while using less labeled data.
    \item We observe the high accuracy of the Louvain algorithm in these clustering tasks. We call attention to leverage community detection approaches in text mining tasks when the true number of clusters is unknown.
    \item We demonstrate the pipeline is scalable with experiments on three different configurations of GPU accelerators. The horizontal and vertical scalability of the pipeline can deal with large datasets and return clustering results to data analysts in a short time.
\end{enumerate}

\section{The Pipeline}
\begin{figure} [h]
\includegraphics[width=1.0\linewidth]{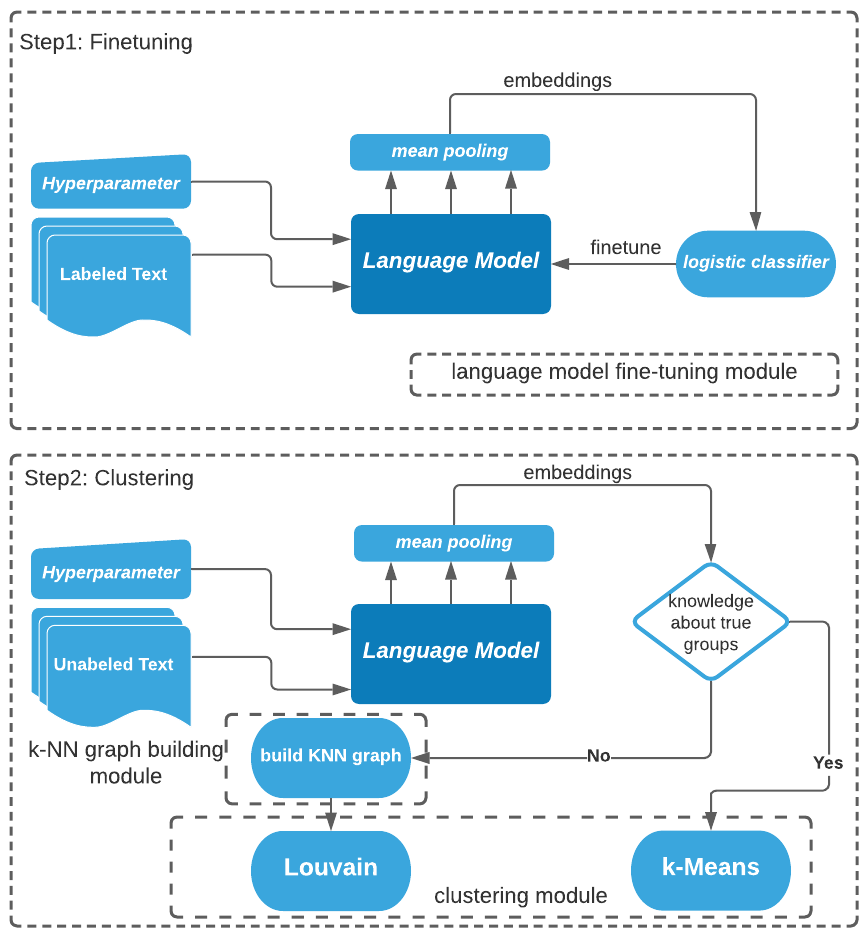}
\caption{ The semi-supervised pipeline for mining intentions from texts. \textbf{Step1:}  Fine-tuning a language model (e.g., BERT). \textbf{Step2:} Use the fine-tuned language model for embeddings. If the true number of clusters is unknown, build k-NN graph and use Louvain; Otherwise directly use k-means to find underlying groups of topics/intentions.}
\label{fig:pipeline}
\end{figure}

Figure~\ref{fig:pipeline} shows the two steps of our pipeline for flexible text clustering.  The core component is fine-tuning and embedding with the language model. The k-NN graph building component and the downstream clustering component serves to surface the topics or intentions groups. In this section, we describe the design and implementation details of each component.

\subsection{Fine-tuning and Embedding with the Language Model}
The first component is the fine-tuning/embedding module. We add a mean pooling layer on top of a language model to get the dense representations for sentences\footnote{\citet{peters2019tune} suggests using a max pooling for the classification tasks, but our experiments show mean pooling consistently gets better clustering quality}. In the initial state when no data has been labelled yet, the untuned language model representations are directly used for downstream clustering as a fully unsupervised approach.  As the analyst works within the VIM interface (see Figure~\ref{fig:prompt}) they begin to organize and label some of the texts from the unsupervised clustering result or from manually created regex patterns.  Once the analyst has provided some small subset of labeled samples it is possible to fine-tune the language model to their particular task.  To fine-tune a language model, we add a linear classifier on top of the mean pooling layer. We optimise the cross entropy loss function to update the language model's parameters and the classifier at the same time.  After this fine-tuning stage, we discard the linear classifier and use the updated language model and mean pooling to produce text representations, i.e., the module maps $N$ input texts or sentences $S=\{s_1,...,s_N\}$ to $N$ dense vectors $X=\{x_1,...,x_N\}$. 

\subsection{Building the k-NN Graph}
The k-nearest neighbors graph is a key data structure to enable the application of a large pool of graph and network analysis methods in data mining~\cite{hajebi2011fast}. Using the \textit{faiss} library~\cite{johnson2019billion}, we convert the dense vector representations $X$ into a k-NN graph. The faiss library can efficiently provide similarity search and clustering with GPU support. On the compressed-domain, the library is able to construct approximate k-NN graphs for up to 1 billion vectors. However, this level of scaling is not anticipated in IVA development. Given the maximum observed volume of input texts of VIM is in the millions of sentences, we can construct the k-NN graph by using exact searching with L2 distances in a few seconds.

\subsection{Downstream Clustering}
We use k-means~\cite{johnson2019billion} or Louvain~\cite{blondel2008fast} depending on if the user specifies the desired number of clusters. Through a sampling mechanism, the faiss library  provides a fast k-means GPU implementation that we incorporate into the downstream clustering component.  The Louvain algorithm can detect hierarchical clustering structures from large networks with millions of nodes and billions of links. Both algorithms can quickly handle the volumes of text commonly worked on by VIM users. The clustering results are presented to analysts (see screenshots in Figure~\ref{fig:squad-auto-cluster}) where they can click on a cluster to see member texts and manually edit cluster membership or launch a sub-clustering task to create intent hierarchies.

\section{Experiment Setup}
\label{sec:evaluation}
In this section, we describe the experiment setup to evaluate and analyze the semi-supervised clustering pipeline\footnote{Datasets and experiment source code: \url{https://nextit-public.s3.us-west-2.amazonaws.com/FinetuneLM-Clustering.zip}}.  

\subsection{Datasets and Tasks}
\label{sec:datasets}

We selected three clustering tasks and preprocessed the text datasets: the Reuters Corpus Volume I~\cite{lewis2004rcv1}, the Python Questions from Stack Overflow~\cite{stackoverflow} and the PubMed 200k RCT~\cite{dernoncourt2017pubmed}. The statistics of selected data and tasks are detailed in Table~\ref{T1}.  For the rest of this paper, we will use their abbreviated names  RCV1, STKOVFL and RCT. We also use the term \textit{topics} or \textit{intentions} when referring to the ground truth classes. 

\begin{table}[t]
\centering
\small
\begin{tabular}{lllr}
\multicolumn{4}{c}{\textbf{Experiment Text Clustering Datasets}} \\
\textbf{} & \textbf{Domain} & \textbf{Classes} & \textbf{Documents} \\
\toprule
RCV1 & News & 4 Topics & 673,771 \\
RCT & Biomedical & 5 Intentions & 270,599 \\
STKOVFL & Programming & 10 Tags & 91,837 \\
\bottomrule
\end{tabular}
\caption{
The task domains, number of classes, and dataset sizes of text datasets used for evaluation of the pipeline. 
}
\label{T1}
\end{table}

We used a train, validate and test paradigm to evaluate the clustering pipeline and get an overview of the required amount of labeled data for the fine-tuning purpose. The train set is used to fine-tune a language model; the test set is embedded and downstream clustering methods are applied. The validation set is kept to help us to evaluate and save the best model during the fine-tuning iterations.   We first select the validation data so that it has the same class distributions as the whole data and include at least 10 samples of each class. We use stratified sampling method to select $800$ posts from STKOVFL; $1,600$ articles from RCV1 and $1,008$ sentences from RCT for validation.   The remaining data of each task was repeatedly split into different sized train and test sets. We started the experiment with $0.0\%$ labeled data (untuned baseline) and increased training set sizes from $0.5\%$, to $1.0\%$, $2.5\%$, $5.0\%$, $10\%$, leaving the rest for test sets. We follow a cross-validation approach to do the stratified sampling 10 times for each training set size. 

The k-means and Louvain algorithms both start from random initialization. We run the two clustering algorithms five times respectively on each of the test set embeddings and report the average clustering quality.  For the untuned baseline experiments, this average is reported over $5$ runs as there was no test/train split. For experiments with fine-tuning steps, the average is over $5$ runs $\times$ $10$ folds $= 50$ runs.

\subsection{Pipeline Parameters}

We select the uncased-large BERT~\cite{devlin2018bert} in PyTorch framework as the language model in the pipeline's core component. The embedded representations are $1024$-dimensional dense vectors.  We conducted preliminary experiments with various other large scale language models and found BERT was the best balance of accuracy and fine-tuning time for our production needs in VIM.  We use the faiss k-means and we implement the Louvain algorithm with C++. All experiments were done on the following Amazon Elastic Compute Cloud (EC2) instances:

\begin{tabular}{llll}
     \textit{Instance} & \textit{vCPUs} & \textit{RAM} & \textit{GPUs} \\
     p3.2xlarge & 4 & 61GB & 1 x Tesla V100 \\
     p3.8xlarge & 32 & 488GB & 4 x Tesla V100 \\
     p3.16xlarge & 64 & 732GB & 8 x Tesla V100 \\
\end{tabular}

We used AdamW optimizer~\cite{loshchilov2017decoupled} and a linear learning rate warm-up scheduler with $10\%$ of the training iterations to fine-tune the language model. For each training set size, we used a sweeping strategy to find the best hyper-parameters based on the F-1 score measured from the validation set. We limit the maximum input sequence length to 128 to avoid out-of-memory issues.

\subsection{Evaluation Metrics}

Clustering Purity and Normalized Mutual Information (NMI) are commonly used metrics for cluster quality evaluation~\cite{manning2008introduction}.  The purity score ranges from $0.0$ to $1.0$. Higher purity means most texts belong to one true class in each detected cluster. Equation~\ref{eq:purity} is for computation of purity score, where 
$\Omega=\{\omega_1,\omega_2,...,\omega_k\}$ is the result clusters, $C=\{c_1,c_2,...,c_j\}$ is the true classes and $N$ is the total number of texts.  The purity score is equivalent to the clustering accuracy (ACC) in other data mining literature.
\begin{equation}\label{eq:purity}
purity(\Omega,C) = \frac{1}{N} \sum_{k} \max_{j}(\omega_{k} \cap c_{j})
\end{equation}

The Normalized Mutual Information (NMI) penalizes splitting data into a large number of small clusters to correct this shortcoming of the purity score. The NMI score ranges from $0.0$ to $1.0$. Higher NMI means good correlation between predicted cluster labels and true classes.  Equation~\ref{eq:nmi} is for computation of NMI score, where $I(\Omega,C)$ is the mutual information, $H(\Omega)$ and $H(C)$ are the entropy. We use both the purity scores and NMI scores to measure downstream clustering qualities.  
\begin{equation}\label{eq:nmi}
NMI(\Omega,C) = \frac{I(\Omega,C)}{[H(\Omega)+H(C)]/2} 
\end{equation}

\section{Analysis And Discussion}
In this section, we analyze the experiment results and try to answer the following questions regarding to our pipeline design: (1) How much labeled data does the pipeline need to fine-tune the language model? (2) How accurately can the pipeline separate different groups of texts? (3) Can Louvain detect high quality clusters without knowing the true number of groups? (4) How fast can the pipeline complete end-to-end clustering tasks on different sized datasets?

\subsection{Amount of Labeled Data For Fine-tuning}

\begin{figure}[h]
\includegraphics[width=\linewidth]{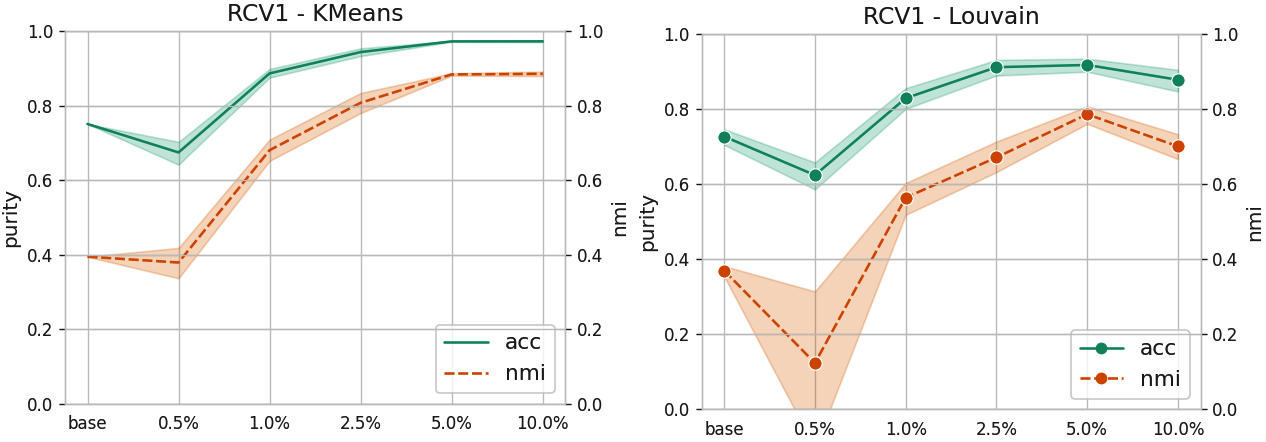}\\
\includegraphics[width=\linewidth]{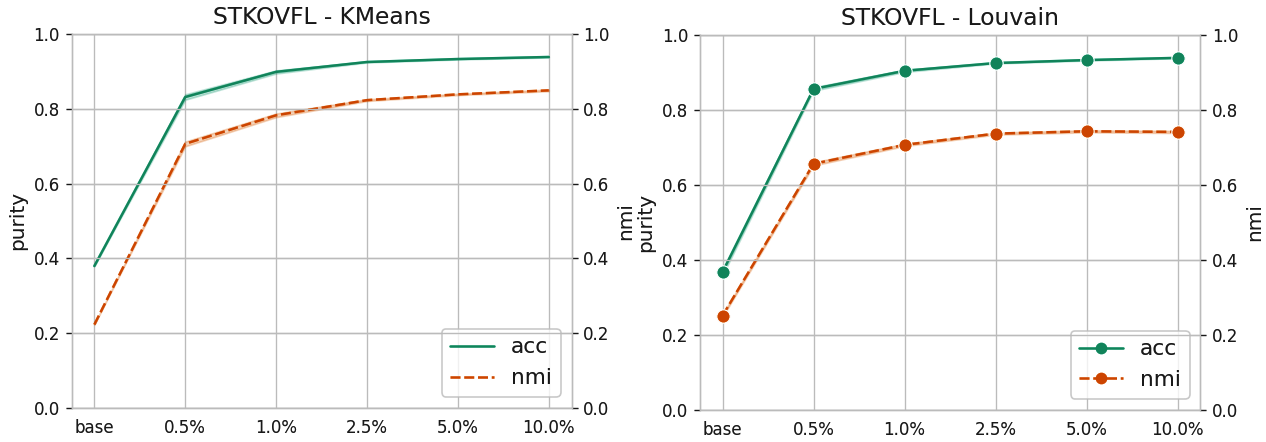}\\
\includegraphics[width=\linewidth]{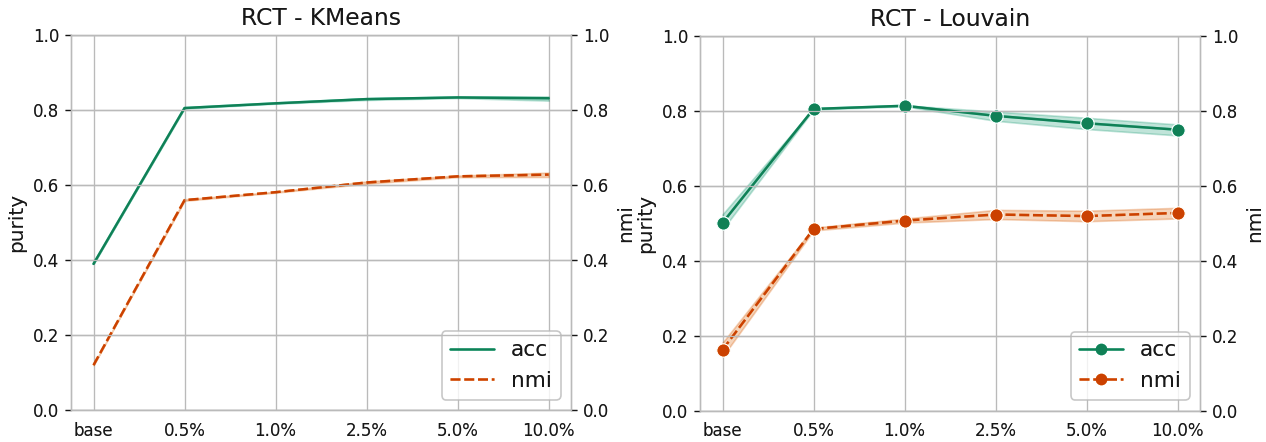}
\caption{ Clustering quality by amount of labeled data used for fine-tuning language models. Left column: Using k-means. Right column: Using Louvain. Solid lines: Purity. Marked Dash lines: NMI.}
\label{fig:kmeans}
\end{figure}

Figure~\ref{fig:kmeans} shows the clustering qualities of the semi-supervised pipeline as we increase the amount of labeled data for fine-tuning.  The left column shows results from using k-means; the right column shows results from using Louvain.  The effect of fine-tuning on downstream clustering quality depends on whether the tasks are from overlapping or novel domains. For tasks that are from similar domains as the language model pre-training corpora (RCV1), the clustering quality first decreases when we use a small amount of labeled texts ($0.5\%$ of the dataset) to fine-tune BERT. This is the known catastrophic forgetting issue that is observed in using transfer learning for classification tasks~\cite{chronopoulou2019embarrassingly}. When we increase the amount of labeled data to $1.0\%$ for fine-tuning, BERT overcomes the forgetting problem and begins to improve downstream clustering quality. 

On the other hand, there is no catastrophic forgetting issue for tasks that are from novel domains (STKOVFL) or when the clustering goal is to find groups of intentions (RCT). The language model needs only a small amount of labeled data ($0.5\%$) to improve the downstream clustering qualities. The clustering qualities of all three tasks reach a plateau after BERT is fine-tuned by only $2.5\%$ labeled data. We determined that using more data for fine-tuning is not beneficial to VIM users because the pre-training time becomes much longer with no performance benefit.  Compared to~\citet{wang2016semi} which paired a CNN or LSTM with k-means and recommends $10\%$ or greater labeled samples to optimize, our approach requires much less labeled data.

\begin{table}[ht]
\small
\begin{tabular}{llcrr}
\multicolumn{5}{c}{\textbf{Clustering Quality Comparison}} \\
\textbf{Task} & \textbf{Method} &  \textbf{\# Clusters} & \textbf{Purity}& \textbf{NMI} \\
\toprule
RCV1 & Baseline+KM  & 4.0 & 0.757 & 0.402  \\
    & Baseline+LV  & 20.2 & 0.789 & 0.441  \\
    & Finetune+KM  & 4.0 & \textbf{0.944} & \textbf{0.808}  \\
    & Finetune+LV  & 20.6 & 0.913 & 0.672  \\
    & DEC  & 4.0 & 0.756 &  N/A  \\
    & DGG$\dagger$  & 4.0 & 0.823 & N/A   \\
    & ss-S$^3$C$^{2}\dagger$  & N/A & N/A & 0.778   \\
 \midrule
STKOVFL & Baseline+KM & 10.0 & 0.346 & 0.189  \\
    & Baseline+LV & 6.0 & 0.433 & 0.328  \\
    & Finetune+KM  & 10.0 & \textbf{0.926} & \textbf{0.823}  \\
    & Finetune+LV  & 35.4 & 0.926 & 0.737  \\
 \midrule
RCT & Baseline+KM  & 5.0 & 0.404 & 0.112  \\
    & Baseline+LV  & 15.0 & 0.611 & 0.252  \\
    & Finetune+KM  & 5.0 & \textbf{0.830} & \textbf{0.608}  \\
    & Finetune+LV  & 10.4 & 0.787 & 0.524   \\
\bottomrule
\end{tabular}
\caption{
After fine-tuning BERT with $2.5\%$ of the data, both downstream clustering methods achieve better clustering quality than the baseline methods. Our pipeline also significantly outperformed existing semi-supervised clustering methods on the RCV1 dataset. {\footnotesize $\dagger$ DGG and ss-S$^3$C$^{2}$ only use a size $10,000$ random sample of the RCV1 data.}
}
\label{T2}
\end{table}

\begin{table}
\centering
\small
\begin{tabular}{llll}
\multicolumn{4}{c}{\textbf{Louvain Clustering Quality Compared to k-means}} \\
\textbf{Task } &  \textbf{\# Clusters} & \textbf{Purity}& \textbf{NMI} \\
\toprule
\textbf{RCV1} & 20.6 (4.0) & 0.913 (0.944) &  0.672 (0.808)  \\
 \midrule
\textbf{STKOVFL} & 35.4 (10.0) & 0.926 (0.926) & 0.737 (0.823)   \\
 \midrule
\textbf{RCT} & 10.4 (5.0) & 0.787 (0.830) & 0.524 (0.608)  \\
\bottomrule
\end{tabular}
\caption{
Clustering quality of Louvain and (k-means) using 2.5\% labeled samples for fine-tuning each dataset.
}
\label{T3}
\end{table}

\subsection{Clustering Accuracy}

In Table~\ref{T2}, for each task, we show the baseline clustering quality using the untuned BERT and the improved accuracy using fine-tuned BERT (with $2.5\%$ labeled samples). For all three tasks, the semi-supervised pipeline improved clustering quality from baseline.  We compared the results with DEC~\cite{xie2016unsupervised}, DGG~\cite{yang2019deep}  and ss-S$^3$C$^2$~\cite{smieja2020classification}, which reported the highest clustering accuracy on the same RCV1 clustering task. DEC and DGG are fully unsupervised approaches whereas ss-S$^3$C$^2$ is a semi-supervised approach.  Our Baseline (unsupervised) is comparable in performance to DEC while our Finetuned pipeline exceeds them all by a large margin.  On the STKOVFL clustering task, our pipeline achieves 0.92 purity score and 0.82 NMI score; on the RCT task, the purity score and NMI score are 0.83 and 0.61 respectively.  The high clustering quality indicates the semi-supervise pipeline can help analysts in both familiar and novel domains. It also shows the benefit of fine-tuning in finding topics, intentions or other clustering groups.

\subsection{Deep Analysis of Louvain and k-means}
In Table~\ref{T3} we compare the clustering quality of Louvain with k-means using the fine-tuned BERT (by $2.5\%$ labeled samples). The number of clusters detected by Louvain is usually greater than the true number of clusters. However, the high purity scores indicates most of the documents within the predicted clusters belong to the same true classes. Considering the documents can have hierarchical topics, one explanation is that Louvain can detect more fine-grained cluster structures inside each high level group from the k-NN graph. 

\begin{table*}[t]
\centering
\small
\begin{tabular}{l|lllll}
\multicolumn{6}{c}{\textbf{Top-10 Frequent Bigrams for the RCV1-GCAT Cluster by Clustering Method}} \\
\toprule
\textbf{K-means} & & & \textbf{Louvain} & \\
\textbf{155,825}    & \textbf{735}& \textbf{21,589} & \textbf{736} & \textbf{98,863} & \textbf{2,047} \\
\midrule
\textbf{prime minister}   & official journal &  press digest & radio romania & prime minister & research institute \\
\textbf{press digest}  & journal contents &  stories vouch & vouch accuracy & hong kong & percent points \\
\textbf{stories vouch}   & contents oj &  verified stories & main headlines & united states & threegrade rating \\
\textbf{verified stories}   & note contents &  reuters verified  & radio headlines & foreign minister & rating system\\
\textbf{reuters verified}   & reverse order &  leading stories  & noon radio & news agency & whose values \\
\textbf{leading stories} & content displayed &  prime minister  & reuter verified & told reuters & issues whose\\
\textbf{hong kong} &  displayed reverse &  vouch accuracy  & afternoon headlines & told reporters & 10 percentage \\
\textbf{united states}   & order printed &  main stories  & romania after & official said &  rating issues \\
\textbf{foreign minister}   & regulation ec &  tuesday reuters  & romania news & human rights & system research \\
\textbf{news agency}   & oj c &  monday reuters  & pm radio & bill clinton & institute assigns \\
& \multicolumn{5}{}{} \\
\textbf{}  &\textbf{1,839} & \textbf{10,698} & \textbf{17,962} & \textbf{4,066} &  \\
\textbf{}  & emergency weather & world cup &  world cup & first innings &  \\
\textbf{ }  & weather conditions & first round &  first division & west indies &  \\
\textbf{}  & conditions wsc & grand prix &  w l & south africa &  \\
\textbf{}  & tropical storm & south africa &  1 0 & sri lanka &  \\
\textbf{}  & top winds & davis cup &  1 1 & new zealand &  \\
\textbf{}  & tropical cyclone & world number &  2 1 & second innings &  \\
\textbf{}  & richter scale & second round &  2 0 & first test &  \\
\textbf{}  & mph winds & number one &  0 0 & second test &  \\
\textbf{}  & 24 hours & czech republic &  0 1 & second day &  \\
\textbf{}  & threat shipping & last year &  result result & lbw b &  \\
\bottomrule
\end{tabular}
\caption{
We compare the top-10 frequent bigrams from fine-grained clusters detected by Louvain with the bigrams from the single GCAT (government and social) cluster found by k-means. The total number of the k-means cluster is $155,825$; the sum of all 9 Louvain clusters is $158,535$. The difference is the true mismatch between k-means and Louvain on this topic category.
}
\label{T4}
\end{table*}

In Table~\ref{T4}, we show the top-10 frequent bigrams from the 9 clusters detected by Louvain and compare them with the bigrams from one big cluster detected by k-means. The true cluster corresponds to the topic category of GCAT (government and social) from the RCV1 data. This example also demonstrates that Louvain is able to detect small groups. For example, $736$ news articles forms a separate group about Romania; $1,839$ articles forms another group about weather. These two clusters consist of only $0.4\%$ and $1.1\%$ of the articles in the entire GCAT category respectively. Another example is the separation of news related to two \textit{world cup} events. By seeing the surfaced bigrams, analysts can understand the first group is more likely about tennis games and the second group is about soccer games.

There are two key takeaways in this observation: first, the k-NN graph accurately connects articles with their nearest neighbors; second, Louvain accurately detects communities with high resolution.  Even though Louvain appears to favor more fine-grained clusters, the comparable clustering quality between the aggregated Louvain clusters and k-means is promising as it means VIM users get similar clustering performance even if they do not know or specify the number of clusters for their task.

\subsection{Computation Time}
The end-to-end computation time for our semi-supervised clustering pipeline consists of the time for fine-tuning, embedding, building k-NN graphs and clustering.  In Table~\ref{T5} we show the  computation time for each step as the proportion in the entire time.  The fine-tuning step and the embedding step are the most time-consuming components. We only report the fine-tuning times using $2.5\%$ labeled samples. If we increase the amount of labeled samples to $5.0\%$ or more, the fine-tuning time and its proportion also increases.  Although we can accelerate the fine-tuning and embedding with more GPUs, the sum of these two steps still contributes to more than $80\%$ of the entire computation time. 

\begin{table}[h]
\footnotesize
\begin{tabular}{p{1.2cm}p{1.3cm}p{0.7cm}p{0.6cm}p{0.5cm}p{0.5cm}p{0.6cm}}
\multicolumn{7}{c}{\textbf{Running Time Breakdown Across Pipeline Stages}} \\
\textbf{Task} & \textbf{Instance} & \textbf{FT} & \textbf{EM} & \textbf{KMS} & \textbf{KNN} & \textbf{LVN} \\
\toprule
RCV1 & p3.2xlarge  & 28.9\% & 64.6\% &  0.1\% & 3.5\% & 2.9\%\\
     & p3.8xlarge  & 27.2\% & 57.8\% &  0.4\% & 3.8\% & 10.8\%\\
     & p3.16xlarge  & 25.1\% & 53.0\% &  1.1\% & 4.4\% & 16.4\%\\
 \midrule
RCT & p3.2xlarge  & 63.6\% & 32.9\% & 0.1\% & 2.1\% & 1.3\%\\
    & p3.8xlarge  & 61.4\% & 30.9\% &  0.9\% & 2.1\% & 4.7\%\\
    & p3.16xlarge  & 58.2\% & 29.1\% &  2.9\% & 2.4\% & 7.4\%\\
 \midrule
STKOVFL & p3.2xlarge  & 49.2\% & 48.5\% & 0.2\%  & 1.3\% & 0.8\% \\
    & p3.8xlarge  & 48.2\% & 46.0\% &  1.6\% & 1.4\% & 2.8\%\\
    & p3.16xlarge  & 47.0\% & 42.0\% &  4.9\% & 2.0\% & 4.1\%\\
\bottomrule
Average & ---  & 45.4\% & 45.0\% & 1.4\%  & 2.6\% & 5.7\% \\    
\bottomrule
\end{tabular}
\caption{
The fine-tuning (FT) and embedding (EM) steps use the majority of the computation time.  This table demonstrates their time can be reduced by adding more GPUs.
}
\label{T5}
\end{table}

\begin{figure}[h]
\hspace*{-0.5cm} 
\includegraphics[scale=0.6]{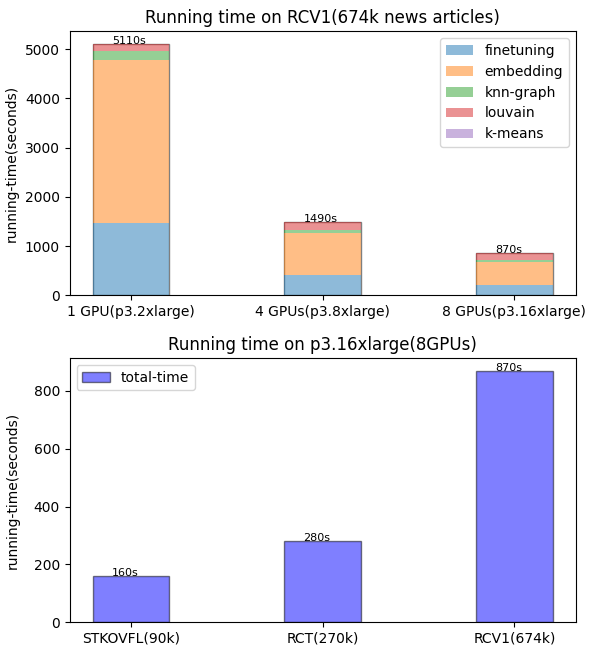}

\caption{The pipeline scales both horizontally and vertically. \textbf{Top}: Break-down of running time of RCV1 data on 3 different Amazon EC2 instances. The pipeline's performance is 3.4x and 5.9x faster than the single GPU instance (p3.2xlarge) when using 4 GPUs (p3.8xlarge) and 8 GPUs (p3.16xlarge) for acceleration. \textbf{Bottom}: The end-to-end time of all 3 datasets on the p3.16xlarge instance. The largest task can be completed within 15 minutes.  }
\label{fig:runtime}
\end{figure}

With the top graph in Figure \ref{fig:runtime}, we show the pipeline is horizontally scalable on more computation resources. The reported running time is tested on 673,000 news articles.  On Amazon EC2 p3.16xlarge instance (8 GPUs), the end-to-end clustering running time is 5.9x faster than on the p3.2xlarge instance (one GPU). 

The bottom graph in Figure \ref{fig:runtime} shows the vertical scalability of the clustering pipeline. Of the three tasks, STKOVFL has over 90,000 sentences, which would be near the average dataset size for the users of VIM that we have observed in production so far. The RCT task resembles a larger dataset size analysts encounter quite often. The RCV1 task contains 0.67 million news articles, which is analogous to a large work load in VIM.  When the pipeline is deployed on Amazon p3.16xlarge instances, the end-to-end computation can be finished under 5 minutes for the daily average tasks, taking closer to 15 minutes for the largest dataset. Moreover, as the clustering pipeline is only used a few times on the full amount of texts during the discovery phase of a new dataset, this performance is acceptable for our application.

\section{System Architecture and Deployment}
VIM is built as a web application using Python and Django for the back-end, Vue.js for the front-end, and PostgreSQL for the database.  The various system components are deployed as Docker containers to Amazon ECS.  The clustering pipeline is implemented as an Amazon SageMaker-compatible Docker container and is deployed as SageMaker model endpoints allowing us to do A-B experiments with different clustering methods and model parameters without changing any of the application code.  The entire system is defined in an Amazon CloudFormation template allowing us to launch the entire application and clustering pipeline service in new production environments quickly.

\section{Application of the Pipeline}
Typical IVA data does not overlap well with the language models' pre-training corpora as they are short texts and usually contain customer specific terminology, therefore BERT must learn to produce task-aware embeddings from fine-tuning. The pipeline enables analysts to iteratively improve their understanding of the input texts data. First, analysts setup a baseline to estimate the number of intention clusters with the untuned language model for embeddings. Analysts can use k-means or Louvain according to their expertise in the text domains. After the first exploration, analysts start to perform some preliminary labeling on a small portion of the data (1 - 5\%) so VIM can fine-tune the language model and improve the clustering quality. In each iteration, VIM remains flexible and configurable to use Louvain or k-means for a variety of datasets and use cases.

After clustering, the results are displayed in the VIM user interface as the counts of high, medium, and low volume clusters and allows users to click on which size clusters they would like to explore.  All three images in Figure~\ref{fig:squad-auto-cluster} below show the result of clicking on the high volume, or largest, clusters.  In this view a box is shown representing each cluster with the size of the box proportionate to the size of the cluster and the number of texts contained in the cluster shown in the top right of the box.  The primary term driving the cluster (nearest the centroid) is shown in the top left of each box.  Within each box is displayed the top terms from that cluster that appeared more often than a configurable threshold.  The font size of each term reflects how often it appeared in relation to the other terms in the cluster, the font color reflects term proximity to each other within the cluster.

Analysts can click on any cluster to see a list of the texts that appeared within that cluster similar to Figure~\ref{fig:prompt}, and can then assign each cluster a human readable label based on the textual contents and also modify the membership of specific texts if they disagree with where the algorithm assigned them.  This allows the analysts to go from a dump of unlabeled text data, to categorized and labeled data very quickly. The analyst can also choose to further refine larger clusters by running the clustering pipeline over just the contents of an existing cluster to create sub-clusters and tease out intent hierarchies. This process can be repeated and each clustering result manually refined until the analyst is satisfied they have flushed out all of the primary intents in the dataset.  We have observed skilled analysts can take a large set of unlabeled texts and produce a large set of high-quality labeled intents and associated texts for downstream training of an IVA in just a few hours using VIM and this iterative approach.

\begin{figure}[h!]
\hspace{-0.25cm}
\includegraphics[scale=0.28]{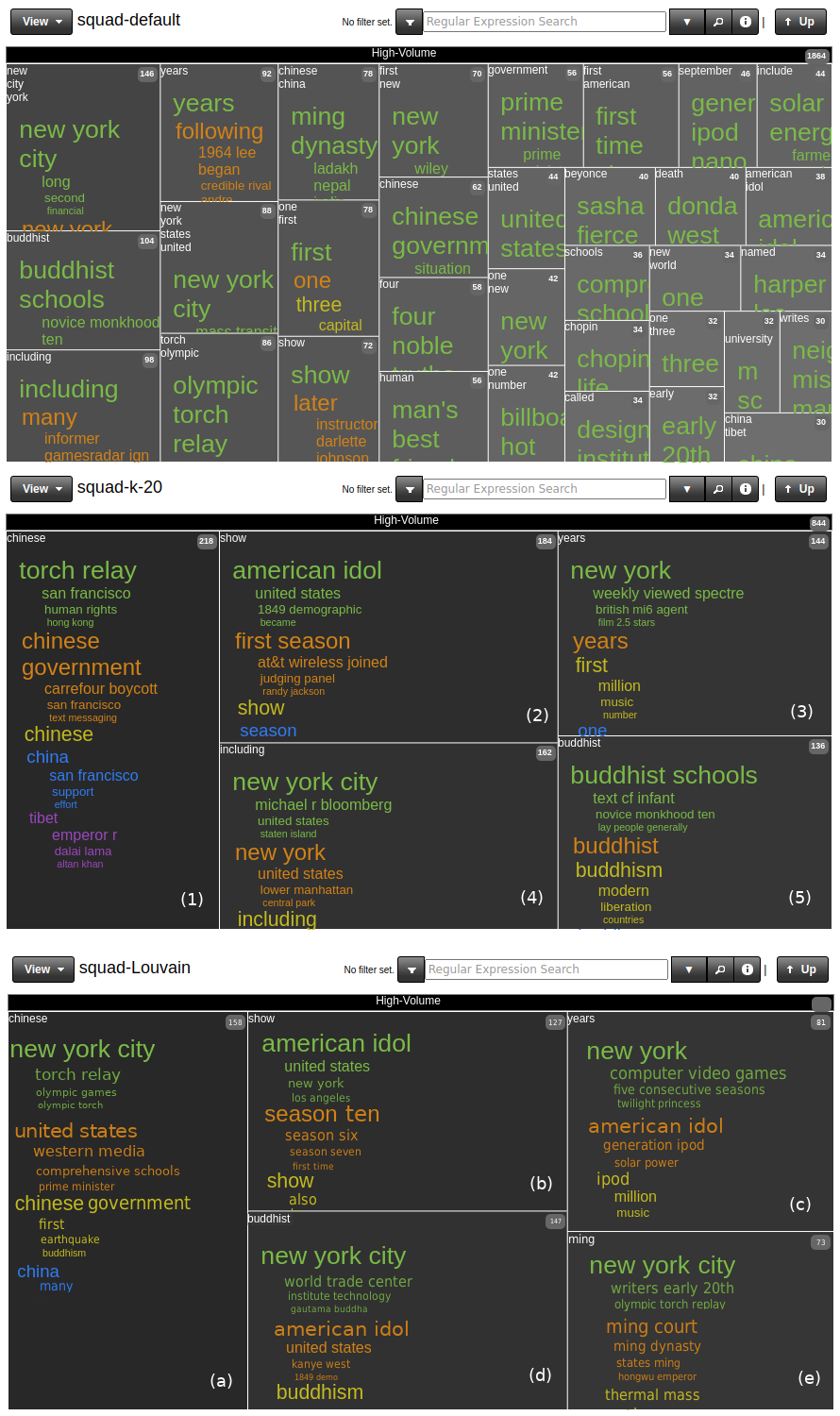}
\caption{VIM's visualization of the frequent terms of clusters from 1,400 paragraphs of Squad2.0 contexts.   \textbf{Top:} The frequent terms displayed to an analyst who uses the default number of clusters $k=250$.  \textbf{Middle:} The frequent terms for an analyst who knows the true number of groups $k=20$. \textbf{Bottom:} The results from Louvain using the untuned BERT for embedding. The number of detected groups $k=14$. }
\label{fig:squad-auto-cluster}
\end{figure}

\section{Case Study}
In previous sections, we demonstrate the usability of our pipeline in terms of it's high clustering accuracy compared with other unsupervised/semi-supervised text clustering methods, but also it's ability to scale. In this section, we use a toy example to demonstrate how our semi-supervised clustering pipeline has improved the usability of the VIM application.  The data consists of 20 randomly selected articles from the SQUAD 2.0 dataset~\cite{rajpurkar2018know} where we extract 1,400 paragraphs.  In this case study, we show how our pipeline provides similar insight to analysts unfamiliar with a dataset as can be discovered by an analyst who is.

There are three scenarios when analysts start working with a customer's input texts.  In the first case, an analyst has no prior knowledge about the data as is the case with a new customer so the potential number of intention groups is unknown.  Before adding our flexible clustering pipeline to the VIM application, in order to explore the data the analyst would either guess $k$ or use the default $k=250$. This default $k$ was historically determined based on the typically larger size of customer datasets analysts work with and in an older version of VIM where k-means was the only clustering option available and it did not leverage pre-trained language models, GPUs, or scale well to large datasets.

In the top panel of Figure~\ref{fig:squad-auto-cluster}, VIM displays the frequent terms selected from the $k=250$ clustering results.
In this example, many frequent terms appear in multiple groups without corresponding contextual terms, such as \textit{one, two, first time, new york}. While some still surface useful topics it is difficult for analysts to review and organize such clusters.

In the second case, the analyst is familiar with the data or a similar business domain, and he or she estimates the number of groups to be $k=20$.  The VIM pipeline completes the clustering task using k-means and displays the frequent terms from 5 high volume clusters in the middle panel of Figure \ref{fig:squad-auto-cluster}. Although there are common frequent terms in multiple clusters, VIM displays other contextual terms and words that were important to each cluster so that analyst can at least identify three meaningful groups. They are cluster$_1$ Chinese government and the Olympic Games, cluster$_2$ American Idol Show and cluster$_5$ Buddhist schools.  

In the bottom panel of Figure~\ref{fig:squad-auto-cluster}, we display the clustering results in the third scenario where an analyst does not provide the number of clusters. Instead, VIM uses Louvain for community detection. With the embeddings from the untuned language model, the pipeline detects 14 groups from the 1,400 paragraphs. The clustering result is not exactly the same as k-means, but the detected groups are a good estimation of the true groups. With this visualization of frequent terms, an analyst can get the clustering result that is analogous to scenario two. In this example, we can still identify the Chinese government and Olympic Games cluster$_a$, the American Idol cluster$_b$ and the Buddhist cluster$_d$. 

\section{Conclusion}
\label{sec:conclusion}

In this paper, we have presented and open-sourced an important AI/ML component in a commercial data exploration and annotation tool used for labeling training data for building IVA intent classifier models.  Our unsupervised and semi-supervised text clustering pipeline is flexible and provides optimal results with a minimum of user configuration (specify number of clusters or not, provided labeled sample for fine-tuning or not). The clustering quality of the pipeline demonstrated considerable improvement over state-of-the-art results on the RCV1 clustering task, and performed well on our other experiments.  The pipeline integrated distributed libraries so that the performance is able to scale vertically on increasing text data volume and horizontally on increasing computation resources. On our current configuration, analysts can get desired clustering results in 15 minutes for the largest task. Our pipeline has greatly reduced the amount of data discovery effort required by the human analysts.

\section{Acknowledgements}
The authors would like to thank Aryn Sargent, Cynthia Freeman, and Dingwen Tao for their feedback and advise.

\bibliography{aaai2021}

\begin{thebibliography}{25}
\providecommand{\natexlab}[1]{#1}
\providecommand{\url}[1]{\texttt{#1}}
\providecommand{\urlprefix}{URL }
\expandafter\ifx\csname urlstyle\endcsname\relax
  \providecommand{\doi}[1]{doi:\discretionary{}{}{}#1}\else
  \providecommand{\doi}{doi:\discretionary{}{}{}\begingroup
  \urlstyle{rm}\Url}\fi

\bibitem[{Aggarwal and Zhai(2012)}]{aggarwal2012survey}
Aggarwal, C.~C.; and Zhai, C. 2012.
\newblock {A survey of text clustering algorithms}.
\newblock In \emph{Mining text data}, 77--128. Springer.

\bibitem[{Beaver and Mueen(2020)}]{beaver2020automated}
Beaver, I.; and Mueen, A. 2020.
\newblock Automated Conversation Review to Surface Virtual Assistant
  Misunderstandings: Reducing Cost and Increasing Privacy.
\newblock \emph{Proceedings of the AAAI Conference on Artificial Intelligence}
  34: 13140--13147.
\newblock \doi{10.1609/aaai.v34i08.7017}.

\bibitem[{Blei(2012)}]{blei2012probabilistic}
Blei, D.~M. 2012.
\newblock Probabilistic topic models.
\newblock \emph{Communications of the ACM} 55(4): 77--84.

\bibitem[{Blondel et~al.(2008)Blondel, Guillaume, Lambiotte, and
  Lefebvre}]{blondel2008fast}
Blondel, V.~D.; Guillaume, J.-L.; Lambiotte, R.; and Lefebvre, E. 2008.
\newblock Fast unfolding of communities in large networks.
\newblock \emph{Journal of statistical mechanics: theory and experiment}
  2008(10): P10008.

\bibitem[{Cer et~al.(2017)Cer, Diab, Agirre, Lopez-Gazpio, and
  Specia}]{cer2017semeval}
Cer, D.; Diab, M.; Agirre, E.; Lopez-Gazpio, I.; and Specia, L. 2017.
\newblock Semeval-2017 task 1: Semantic textual similarity-multilingual and
  cross-lingual focused evaluation.
\newblock \emph{arXiv preprint arXiv:1708.00055} .

\bibitem[{Chronopoulou, Baziotis, and
  Potamianos(2019)}]{chronopoulou2019embarrassingly}
Chronopoulou, A.; Baziotis, C.; and Potamianos, A. 2019.
\newblock An embarrassingly simple approach for transfer learning from
  pretrained language models.
\newblock \emph{arXiv preprint arXiv:1902.10547} .

\bibitem[{Dernoncourt and Lee(2017)}]{dernoncourt2017pubmed}
Dernoncourt, F.; and Lee, J.~Y. 2017.
\newblock Pubmed 200k rct: a dataset for sequential sentence classification in
  medical abstracts.
\newblock \emph{arXiv preprint arXiv:1710.06071} .

\bibitem[{Devlin et~al.(2018)Devlin, Chang, Lee, and
  Toutanova}]{devlin2018bert}
Devlin, J.; Chang, M.-W.; Lee, K.; and Toutanova, K. 2018.
\newblock Bert: Pre-training of deep bidirectional transformers for language
  understanding.
\newblock \emph{arXiv preprint arXiv:1810.04805} .

\bibitem[{Hajebi et~al.(2011)Hajebi, Abbasi-Yadkori, Shahbazi, and
  Zhang}]{hajebi2011fast}
Hajebi, K.; Abbasi-Yadkori, Y.; Shahbazi, H.; and Zhang, H. 2011.
\newblock Fast approximate nearest-neighbor search with k-nearest neighbor
  graph.
\newblock In \emph{Twenty-Second International Joint Conference on Artificial
  Intelligence}.

\bibitem[{Johnson, Douze, and J{\'e}gou(2019)}]{johnson2019billion}
Johnson, J.; Douze, M.; and J{\'e}gou, H. 2019.
\newblock Billion-scale similarity search with GPUs.
\newblock \emph{IEEE Transactions on Big Data} .

\bibitem[{Lewis et~al.(2004)Lewis, Yang, Rose, and Li}]{lewis2004rcv1}
Lewis, D.~D.; Yang, Y.; Rose, T.~G.; and Li, F. 2004.
\newblock R{C}{V}1: A new benchmark collection for text categorization
  research.
\newblock \emph{Journal of Machine Learning Research} 5(Apr): 361--397.

\bibitem[{Liu et~al.(2019)Liu, Ott, Goyal, Du, Joshi, Chen, Levy, Lewis,
  Zettlemoyer, and Stoyanov}]{liu2019roberta}
Liu, Y.; Ott, M.; Goyal, N.; Du, J.; Joshi, M.; Chen, D.; Levy, O.; Lewis, M.;
  Zettlemoyer, L.; and Stoyanov, V. 2019.
\newblock Roberta: A robustly optimized bert pretraining approach.
\newblock \emph{arXiv preprint arXiv:1907.11692} .

\bibitem[{Loshchilov and Hutter(2017)}]{loshchilov2017decoupled}
Loshchilov, I.; and Hutter, F. 2017.
\newblock Decoupled weight decay regularization.
\newblock \emph{arXiv preprint arXiv:1711.05101} .

\bibitem[{Manning, Raghavan, and Sch{\"u}tze(2008)}]{manning2008introduction}
Manning, C.~D.; Raghavan, P.; and Sch{\"u}tze, H. 2008.
\newblock \emph{Introduction to information retrieval}.
\newblock Cambridge university press.

\bibitem[{Peters, Ruder, and Smith(2019)}]{peters2019tune}
Peters, M.; Ruder, S.; and Smith, N.~A. 2019.
\newblock To tune or not to tune? adapting pretrained representations to
  diverse tasks.
\newblock \emph{arXiv preprint arXiv:1903.05987} .

\bibitem[{Peters et~al.(2018)Peters, Neumann, Iyyer, Gardner, Clark, Lee, and
  Zettlemoyer}]{peters2018deep}
Peters, M.~E.; Neumann, M.; Iyyer, M.; Gardner, M.; Clark, C.; Lee, K.; and
  Zettlemoyer, L. 2018.
\newblock Deep contextualized word representations.
\newblock \emph{arXiv preprint arXiv:1802.05365} .

\bibitem[{Radford et~al.(2019)Radford, Wu, Child, Luan, Amodei, and
  Sutskever}]{radford2019language}
Radford, A.; Wu, J.; Child, R.; Luan, D.; Amodei, D.; and Sutskever, I. 2019.
\newblock Language models are unsupervised multitask learners.
\newblock \emph{OpenAI Blog} 1(8): 9.

\bibitem[{Rajpurkar, Jia, and Liang(2018)}]{rajpurkar2018know}
Rajpurkar, P.; Jia, R.; and Liang, P. 2018.
\newblock Know what you don't know: Unanswerable questions for SQuAD.
\newblock \emph{arXiv preprint arXiv:1806.03822} .

\bibitem[{Ram et~al.(2018)Ram, Prasad, Khatri, Venkatesh, Gabriel, Liu, Nunn,
  Hedayatnia, Cheng, Nagar et~al.}]{ram2018conversational}
Ram, A.; Prasad, R.; Khatri, C.; Venkatesh, A.; Gabriel, R.; Liu, Q.; Nunn, J.;
  Hedayatnia, B.; Cheng, M.; Nagar, A.; et~al. 2018.
\newblock Conversational ai: The science behind the alexa prize.
\newblock \emph{arXiv preprint arXiv:1801.03604} .

\bibitem[{{\'S}mieja, Struski, and Figueiredo(2020)}]{smieja2020classification}
{\'S}mieja, M.; Struski, {\L}.; and Figueiredo, M.~A. 2020.
\newblock A classification-based approach to semi-supervised clustering with
  pairwise constraints.
\newblock \emph{Neural Networks} .

\bibitem[{Stackoverflow(2019)}]{stackoverflow}
Stackoverflow, K. 2019.
\newblock Full text of {S}tack {O}verflow {Q}\&{A} about the {P}ython
  programming language.

\bibitem[{Steyvers and Griffiths(2007)}]{steyvers2007probabilistic}
Steyvers, M.; and Griffiths, T. 2007.
\newblock Probabilistic topic models.
\newblock \emph{Handbook of latent semantic analysis} 427(7): 424--440.

\bibitem[{Wang, Mi, and Ittycheriah(2016)}]{wang2016semi}
Wang, Z.; Mi, H.; and Ittycheriah, A. 2016.
\newblock Semi-supervised Clustering for Short Text via Deep Representation
  Learning.
\newblock \emph{CoNLL 2016} 31.

\bibitem[{Xie, Girshick, and Farhadi(2016)}]{xie2016unsupervised}
Xie, J.; Girshick, R.; and Farhadi, A. 2016.
\newblock Unsupervised deep embedding for clustering analysis.
\newblock In \emph{International {C}onference on {M}achine {L}earning},
  478--487.

\bibitem[{Yang et~al.(2019)Yang, Cheung, Li, and Fang}]{yang2019deep}
Yang, L.; Cheung, N.-M.; Li, J.; and Fang, J. 2019.
\newblock Deep clustering by gaussian mixture variational autoencoders with
  graph embedding.
\newblock In \emph{Proceedings of the IEEE International Conference on Computer
  Vision}, 6440--6449.

\end{thebibliography}
\end{document}